\documentclass{article}

\PassOptionsToPackage{numbers, compress}{natbib}
\usepackage[final]{tackling_climate_workshop_style}

\usepackage[utf8]{inputenc} % allow utf-8 input
\usepackage[T1]{fontenc}    % use 8-bit T1 fonts
\usepackage{hyperref}       % hyperlinks
\hypersetup{hidelinks}
\usepackage{url}            % simple URL typesetting
\usepackage{booktabs}       % professional-quality tables
\usepackage{amsfonts}       % blackboard math symbols
\usepackage{nicefrac}       % compact symbols for 1/2, etc.
\usepackage{microtype}      % microtypography

\usepackage{multirow}
\usepackage{enumitem}
\usepackage{amsmath}
\usepackage{graphicx}
\usepackage{svg}
\usepackage{pifont}

\usepackage{algorithm}
\usepackage{algpseudocode}
\usepackage{subcaption}
\usepackage{booktabs}

\usepackage{lscape} 

\usepackage{siunitx} 
\title{Nowcasting of Tropical Cyclone Wind Fields and Intensity from Geostationary Imagery}

\author{%
\parbox{0.95\textwidth}{
\centering
Simon Donike$^{1,2}$ \quad Pritthijit Nath$^{1,3}$ \quad Cristina Radin$^{1,4}$ \\ \vspace{0.1cm} Arthur Avenas$^{1,5}$ \quad Ruben Cartuyvels$^{1,5}$ \quad Emiliano Diaz$^{1,2}$ \\ \vspace{0.1cm}  \quad William K. Jones$^{1,5}$ \quad Anna Jungbluth$^{1,6}$} \vspace{0.2cm} \\
$^1$ Earth Systems Lab, Frontier Development Lab\\
$^2$ University of Valencia, Spain\\
$^3$ University of Cambridge, United Kingdom\\
$^4$ Max Planck Institute for Marine Microbiology, Germany \\
$^5$ European Space Agency \\
$^6$ Imperial College London, United Kingdom}

\newif\ifshowcontent
\newif\ifnotshowcontent

\notshowcontenttrue

\newcommand{\cmark}{\ding{51}}
\newcommand{\xmark}{\ding{55}}

\begin{document}
\setcitestyle{square}

\maketitle
\vspace{-7mm}
\enlargethispage{2\baselineskip}

\begin{abstract}
High-resolution tropical cyclone surface wind speeds derived from synthetic aperture radar (SAR) observations can support storm structure and intensity modelling, but remain spatially and temporally sparse and unavailable near-real-time. Geostationary satellites capture cloud tops continuously in space and time, but do not directly resolve surface winds. 
Here, we present simple neural networks to predict 2D wind fields and storm-scale intensity metrics from geostationary imagery. Through a series of ablation studies, we find that joint prediction of 2D fields and scalar metrics from the model latent space improves the retrieval of the (radius of) maximum wind, especially for rapid intensification events. 
\end{abstract}
% \ifshowcontent{
% \url{insert-github-link}.}\fi
% \ifnotshowcontent{\url{insert-anonymous-github-link-using-anonymous.4open.science}.}\fi
% \end{abstract}

\section{Introduction}
Tropical cyclones (TCs) are destructive natural hazards that form over the warm tropical oceans. Accurately predicting their intensity and evolution remains a challenge, especially during rapid intensification events (RI, at least $30\,\mathrm{kt}$ increase in 24 hours), which leave limited time for evacuation and disaster response~\cite{kaplan_rapidly_2003}. While numerical weather prediction and data-driven forecasting have made progress in predicting TC trajectories~\cite{cangialosi_intensity_2020}, intensity estimates still need to be improved. Moreover, existing models largely focus on scalar outputs, such as maximum sustained wind speed ($V_{\max}$), even though destructive potential is linked to the full wind field through integrated kinetic energy~\cite{powell_tropical_2007}.

Spaceborne synthetic aperture radar (SAR) signal can penetrate cloud cover to provide surface wind speed estimates at kilometre resolution~\cite{mouche_sar_2019}, but overpasses are infrequent, roughly twice a day across the TC lifetime~\cite{avenas_cyclobs_2025}, making it unsuitable for continuous monitoring. Geostationary (GEO) satellite observations, in contrast, provide near-continuous imagery that primarily captures the cloud-top structure, also at a high spatial resolution. GEO has long served as the predominant observational input for data-driven TC intensity estimation, from the original Dvorak technique~\cite{dvorak_satellite_1975,dvorak_satellite_1984} to convolutional neural networks (CNNs) trained to predict maximum wind speed~\cite{chen_rotationblended_2018,maskey_deepti_2020,wang_geostationary_2022}, category~\cite{pradhan_intensity_2018}, or radius~\cite{zhuo_physicsaugmented_2021,wang_physical_2026}, as well as RI forecasting~\cite{griffin_rapidintensification_2022,chen_ensemble_2023,wang_contrastive_2025}. Modelling of high-resolution spatial fields is less common, with previous approaches predicting aircraft-level winds~\cite{wimmers_windretrieval_2024} or downscaling coarse wind products with SAR-derived observations~\cite{zhang_multisource_2025}. These methods, however, either target scalar intensity, reconstruct wind fields at flight level rather than the surface, or refine existing data sets. 

To the best of our knowledge, no prior work has used only satellite-based observations as input to predict high-resolution SAR wind fields \emph{and} storm-scale intensity metrics. Here, we show -- in a nowcasting sertup -- that estimating $V_{\max}$ and wind-field structure during RI events benefits from jointly learning to reconstruct SAR-derived surface wind fields rather than predicting scalar quantities alone. We further show, through a series of ablation studies, that observation-only models are not meaningfully improved by adding coarse reanalysis products as input. With this, our work yields an observation-only workflow suitable for near-real-time downstream applications.

%\clearpage

\section{Data}

\textbf{TC tracks: } IBTrACS~\cite{knapp_ibtracs_2010} compiles historical best-track estimates, including location, intensity, and extent, for TCs. In this work, we use the following metrics: $V_{\max}$ (maximum sustained wind speed), $R_{\max}$ (radius of maximum wind), and $R_{34}$, $R_{50}$, and $R_{64}$ (radii of $34$, $50$, and $64\,\mathrm{kt}$ winds in each geographical quadrant).

\textbf{SAR wind fields: } Surface wind speed estimates from SAR provide target data for spatially resolved fields. These wind retrievals are based on variations in the SAR radar backscatter caused for instance by wind-driven capillary waves on the ocean surface. SAR data are sourced from the CyclObs database~\cite{avenas_cyclobs_2025}, which provides TC surface wind retrievals for a wide variety of SAR instruments at $3\,\mathrm{km}$ resolution.

\textbf{Geostationary satellite imagery: } Imagery from GOES-16 and -18 as well as Himawari-8 and -9 (see Appendix Figure \ref{fig:field-of-view}) provide infrared (IR) brightness temperature measurements across ten shared thermal IR channels at high spatial ($2\,\mathrm{km}$ at nadir) and temporal (10 minute) resolution~\cite{schmit2017closer, bessho2016introduction}.

\textbf{Reanalysis data: } 
As optional inputs, we use seven ERA5~\cite{hersbach_era5_2020} variables (precipitable water, sea-surface temperature, mean sea-level pressure, $2\,\mathrm{m}$ air and dew-point temperature, and $10\,\mathrm{m}$ u- and v-wind) and derive $10\,\mathrm{m}$ wind speed and relative vorticity.

\textbf{Data processing: } We pair CyclObs 2D wind fields with spatially aligned GEO imagery and temporally interpolated IBTrACS metrics, using the closest GEO observation within $\pm90$ minutes of each SAR acquisition to maximize the number of samples. In most cases, this matching results in GEO imagery within $\pm15$ mins from the corresponding SAR observation.

GEO imagery is cropped to $196\times196$ pixels while varying the TC-centre position to reduce spatial bias. Storms are classified as no RI or RI based on 24-hour wind. We perform a stratified 60/20/20 train/validation/test split, keeping observations from the same storm together to prevent data leakage. This yields 841/232/212 samples from 115/34/38 storms, including 134 RI samples from 62 storms.

\section{Models}

\begin{figure}[!h]
    \centering
    \includegraphics[width=\linewidth]{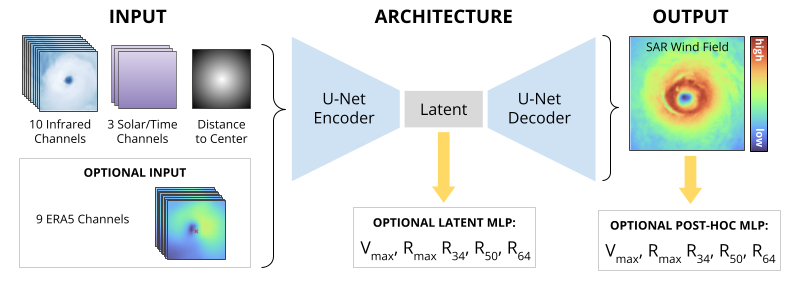}
    \caption{Overview of our modelling pipeline. GEO imagery, spatial context channels, and optionally ERA5 are translated into SAR-derived wind fields via a U-Net. An MLP can be added to the latent space, or after the 2D field output, to predict the maximum and radius of maximum wind ($V_{\max}$, $R_{\max}$), as well as the radii of 34, 50, and 64 $\mathrm{kt}$ winds ($R_{34}$, $R_{50}$, $R_{64}$).}
    \label{fig:pipeline-diagram}
\end{figure}

Figure~\ref{fig:pipeline-diagram} shows the modelling pipeline. The input contains ten GEO channels and four spatial context channels: local-solar-time sine/cosine, solar zenith angle, and distance to storm centre. Experiments optionally add nine ERA5 channels, and a validity mask is appended to the resulting 14- or 23-channel input. Here, $I_{\mathrm{GEO}}$ denotes the complete input tensor, with evaluation against $V_{\max}$ and $R_{\max}$.

As a baseline, a CNN encoder computes $z=\text{enc}_{\theta}(I_{\mathrm{GEO}})$ without training against the 2D wind field. Here, an MLP maps global mean- and maximum-pooled features to $(\hat V_{\max},\hat R_{\max})=\text{MLP}_{\phi}(z)$. We compare the baseline with three wind field models:

%\clearpage
\enlargethispage{\baselineskip}

\begin{enumerate}
    \item \textbf{U-Net: } A U-Net~\cite{ronneberger_unet_2015} predicts $\hat I_{\mathrm{Wind}}=\text{dec}_{\eta} (\text{enc}_{\theta}(I_{\mathrm{GEO}}))$. With ERA5 input, the model predicts a residual added to the ERA5 field, otherwise, it predicts absolute wind. From these predicted wind fields, we derive $\hat V_{\max}$ and $\hat R_{\max}$ and compare them with the IBTrACS metrics.  
    
    \item \textbf{Post-hoc MLP: } A two-stage model uses a frozen U-Net and predicts $(\hat V_{\max},\hat R_{\max})=\text{MLP}_{\zeta} (\text{U-Net}_{\psi}(\hat I_{\mathrm{Wind}}),m)$, where the U-Net encodes the field, validity mask, and storm-centred distance, and $m$ contains current-time and location metadata.
    
    \item \textbf{Latent MLP: } A joint model shares the features $z=\text{enc}_{\theta}(I_{\mathrm{GEO}})$ between
    $\hat I_{\mathrm{Wind}}=\text{dec}_{\eta}(z)$ and $(\hat V_{\max},\hat R_{\max})=\text{MLP}_{\phi}(z)$.
\end{enumerate}

All U-Nets are trained with loss $\mathcal{L}_{\mathrm{Wind}}$, which upweights winds between $25$ and $50\,\mathrm{m\,s^{-1}}$ on a scale from $1\times$ to $8\times$ and includes a $0.05$-weighted peak of the top 0.5 percentile values. All MLPs are trained with $\mathcal L_{V_{\max}} +0.25~\mathcal L_{R_{\max}}$, using storm- and intensity-balanced weights. We also compare scalar supervision of only $V_{\max}$ and $R_{\max}$, with models additionally trained against $R_{34}$, $R_{50}$, $R_{64}$ for more spatial nuance. SAR-derived wind fields are training targets only and are therefore not required at inference time. All objectives use Huber losses in physical units. The Huber transitions are $2\,\mathrm{m\,s^{-1}}$, $5\,\mathrm{m\,s^{-1}}$, and $20\,\mathrm{km}$ for the wind field, $V_{\max}$, and $R_{\max}$, respectively.
Further hyperparameter settings are provided in \ref{app:hyperparams}.

\begin{table*}[hb]
    \centering
    \caption{(\subref{tab:vmax-architecture-ablation}) $V_{\max}$ MAE ($\mathrm{m\,s^{-1}}$) on the held-out test set. The U-Net only predicts the wind field, all other models additionally predict $V_{\max}$, $R_{max}$, $R_{34}$, $R_{50}$, $R_{64}$. RI denotes rapid-intensification observations, bold values are best within
    each training regime. 
    (\subref{tab:latent-supervision-ablation}) MAE for the joint U-Net and latent MLP supervision ablations, trained to predict the wind fields and either just $V_{\max}$ or also $R_{max}$, $R_{34}$, $R_{50}$, $R_{64}$. $V_{\max}$ is
    reported in $\mathrm{m\,s^{-1}}$ and $R_{\max}$ in $\mathrm{km}$. Note that these metrics were calculated over a reduced test set where the TC centre falls on a valid SAR pixel, as required for radii calculations, lowering the number of storms from 38 to 33. Bold values are the best in each column.}
    \begin{subtable}{0.45\textwidth}
    \centering
    \footnotesize
    \setlength{\tabcolsep}{4pt}
    \caption{}
    \begin{tabular}{lcrr}
        \toprule
        Model & ERA5 & All & RI \\
        \midrule
        U-Net & \ding{55} & 7.518 & 17.859 \\
        Post-hoc MLP & \xmark & 6.715 & 12.507 \\
        Latent MLP & \xmark & \textbf{5.885} & \textbf{8.934} \\
        \midrule
        U-Net & \cmark & 6.446 & 14.475 \\
        Post-hoc MLP & \cmark & \textbf{5.529} & 10.512 \\
        Latent MLP & \cmark & 7.302 & \textbf{7.716} \\
        \bottomrule
    \end{tabular}    
    \label{tab:vmax-architecture-ablation}
    \end{subtable}\begin{subtable}{0.55\textwidth}
    \centering
    \footnotesize
    \setlength{\tabcolsep}{3pt}
    \caption{}
    \begin{tabular}{ccrrrr}
        \toprule
        & & \multicolumn{2}{c}{$V_{\max}$} &
        \multicolumn{2}{c}{$R_{\max}$} \\
        \cmidrule(lr){3-4}\cmidrule(lr){5-6}
        SAR & Radius & All & RI & All & RI \\
        \midrule
        \cmark & \xmark  & 6.694 & 6.867 & -- & -- \\
        \xmark  & \cmark & \textbf{6.297} & 7.198 & 19.44 & 13.14 \\
        \xmark  & \xmark  & 6.828 & 6.645 & -- & -- \\
        \cmark & \cmark & 6.344 & \textbf{5.469} & \textbf{18.24} & \textbf{11.18} \\       
        \bottomrule
    \end{tabular}
    
    \label{tab:latent-supervision-ablation}
    \end{subtable}    
\end{table*}

\section{Results}

We report $V_{\max}$ and $R_{\max}$, both over all observations and RI.

\textbf{Architecture and ERA5 ablations: } Table \ref{tab:vmax-architecture-ablation} shows the comparison across different training strategies and model inputs. Without ERA5, the U-Net and latent MLP  performs best, with a $V_{\max}$ MAE of $5.885\,\mathrm{m\,s^{-1}}$, compared with $6.715$ for the post-hoc MLP and $7.518\,\mathrm{m\,s^{-1}}$ for the simple U-Net without a scalar-guided loss. It also performs best during RI ($8.934$ vs. $12.507$ and $17.859\,\mathrm{m\,s^{-1}}$). ERA5 improves the U-Net and post-hoc MLP, and also reduces the error for RI with the latent MLP, although the error is not reduced when considering all storms. 

\textbf{Scalar supervision ablations: } For the GEO-only latent MLP model (see  \autoref{tab:latent-supervision-ablation}), combining SAR wind field reconstruction with radius supervision reduces RI $V_{\max}$ MAE from $7.198$ to $5.469\,\mathrm{m\,s^{-1}}$ and RI $R_{\max}$ MAE from $13.14$ to $11.18\,\mathrm{km}$. Without radius supervision, SAR alone does not improve RI $V_{\max}$, indicating that the spatial and scalar objectives are complementary. 

%\clearpage
%\vspace*{-1cm}
%\enlargethispage{2\baselineskip}

\begin{figure}[ht]
    \centering
    \includegraphics[width=0.95\textwidth, 
    trim=0cm 0cm 0cm 0.2cm]
    {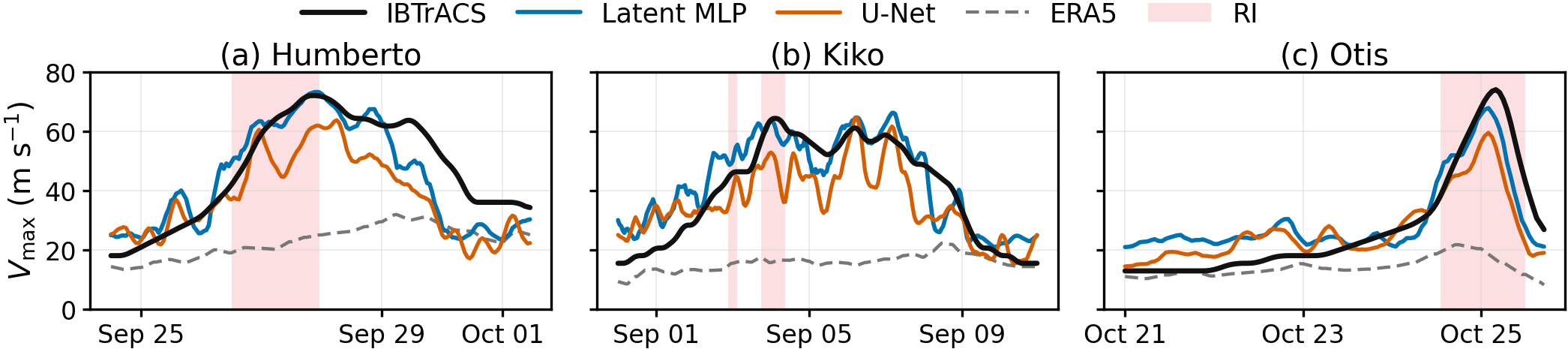}
    \caption{Complete-storm $V_{\max}$ nowcasts from the latent MLP (model 3) and field-only
    U-Net (model 1), both without ERA5 input. Shading denotes RI periods. Curves use
    a centred five-hour rolling mean, while metrics use native predictions.}
    \label{fig:full-storm-nowcasts}
\end{figure}

\textbf{Complete-storm case studies: } Figure~\ref{fig:full-storm-nowcasts} compares the GEO-only U-Net and latent MLP models over the complete life cycles of Hurricane Humberto (North Atlantic, 2019), Kiko (East Pacific, 2025), and Otis (East Pacific, 2023). Across the 3,266 observations of these storms, the latent MLP reduces $V_{\max}$ MAE from $9.197$ to $7.434\,\mathrm{m\,s^{-1}}$ relative to the simple U-Net, and from $12.203$ to $5.441\,\mathrm{m\,s^{-1}}$ during RI. 
 
\begin{figure}[H]
    \centering
    \includegraphics[width=0.80\textwidth, trim=0.5cm 0.2cm 0.5cm 1cm]
    {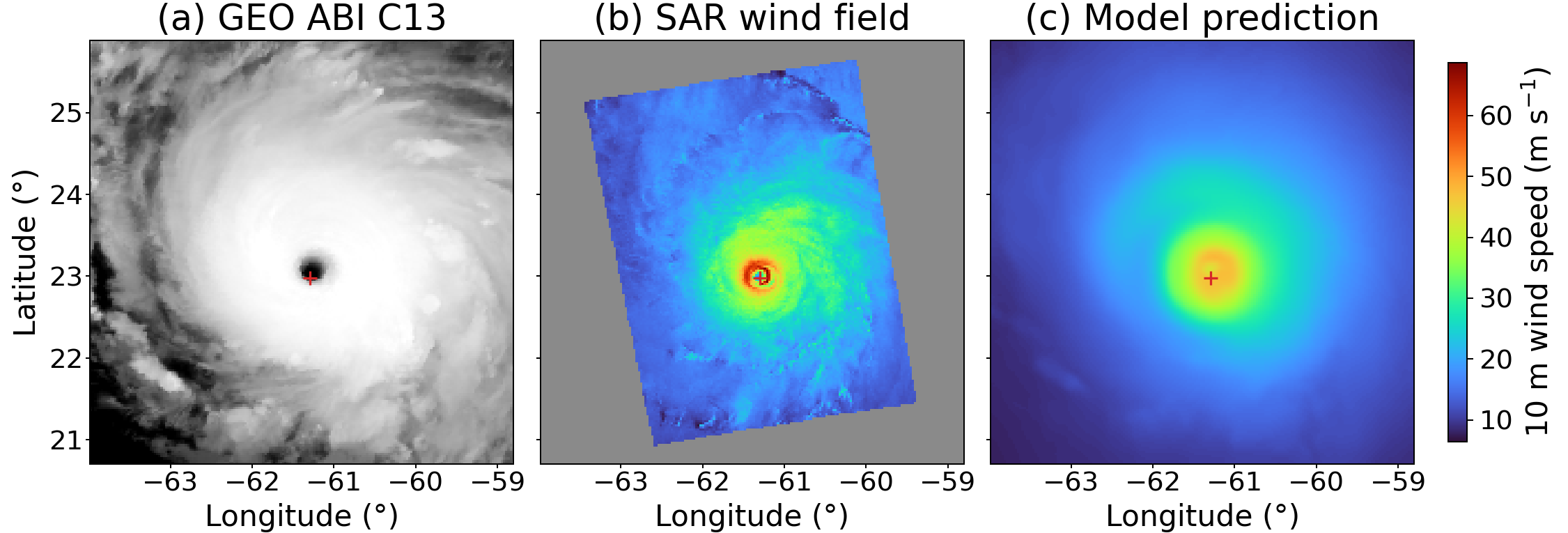}
    \caption{Wind field reconstruction of a category 5 storm. (a) Geostationary imagery (infrared channel 13), (b) target SAR-derived wind field, and (c) wind field prediction from the latent MLP model. Red crosses mark the IBTrACS storm centre.}
    \label{fig:sar}
    \vspace{-2mm}
\end{figure}

\autoref{fig:sar} shows a wind field reconstruction example at category 5 intensity. While the output is smoothed compared with the SAR-derived reference field, we correctly predict the extent of winds $\geq 30\,\mathrm{m\,s^{-1}}$ and can fill in beyond the edge of the SAR swath.

\section{Conclusion}
In this work, we derive 2D wind fields and scalar metrics of TCs through their life cycles, including RI. Our best model jointly optimises a U-Net for field prediction with a latent MLP to predict storm-scale quantities such as the (radius of) maximum sustained wind. Jointly supervising the latent representation with SAR wind fields and storm-scale quantities is particularly beneficial during RI, when accurately representing fine-scale wind structure is most important. We show that both the wind fields and metrics can be derived from geostationary imagery alone.

While ERA5 conditioning provides useful environmental context and improves performance in some configurations, removing it moves the workflow away from reanalysis-based, non-observational inputs. Our resulting model requires only near-real-time GEO imagery and deterministic spatial context at inference, improving its availability and usability for operational monitoring and other downstream applications. Initial exploration of more complex, higher-parameter models~\cite{chen_deeplab_2018,zhou_unetplusplus_2020,xie_segformer_2021,dauvilliers_motif_2025} did not outperform our approach, largely due to the limited amount, quality, and balance of training data.

We therefore opted for simple ML approaches that allow for quick experimentation and easy ablation studies. Future work will explore alternative architectures, for instance based on vision transformers, which have shown great promise for geospatial analysis. Beyond limitations arising from the training data and the need for paired examples, we acknowledge several areas for future work. We have not yet conducted extensive ablation studies of the ERA5 inputs to determine whether alternative combinations of input channels could further improve performance, nor have we explored additional hyperparameter tuning, including the choice of loss function. Additional geophysical variables, e.g., describing the upper ocean conditions along the forecasted track, may further improve our estimates. Nevertheless, the ability to recover both spatial wind structure and storm-scale intensity metrics from GEO imagery alone provides a promising foundation for observation-driven TC monitoring and potential future integration into near-real-time monitoring workflows.

\ifshowcontent{

}\fi

\section*{Code Availability}
The code developed for this research study along with the StormSense dashboard is publicly available at \url{https://bit.ly/3WRhXqx}.

\begin{ack}
This work has been enabled by Frontier Development Lab Earth Systems Lab (\url{https://eslab.ai/}) a public-private partnership between the European Space Agency (ESA), Trillium Technologies and University of Oxford and leaders in commercial AI supported by Scan Computers, ESA Space HPC, UK AI Research Resource (AIRR), JASMIN, Google Cloud and NVIDIA Corporation. 
\end{ack}

\small
\bibliographystyle{vancouver}
\bibliography{bibliography}

\clearpage
\normalsize
\appendix
\renewcommand\thesection{Appendix \Alph{section}}
\renewcommand\thesubsection{\Alph{section}.\arabic{subsection}} 
\renewcommand\thetable{\Alph{section}.\arabic{table}}  
\renewcommand\thefigure{\Alph{section}.\arabic{figure}}  
\setcounter{table}{0}
\setcounter{figure}{0}

\section{Geostationary Field-of-Views}

\begin{figure}[htb]
    \centering
    \includegraphics[width=\linewidth]{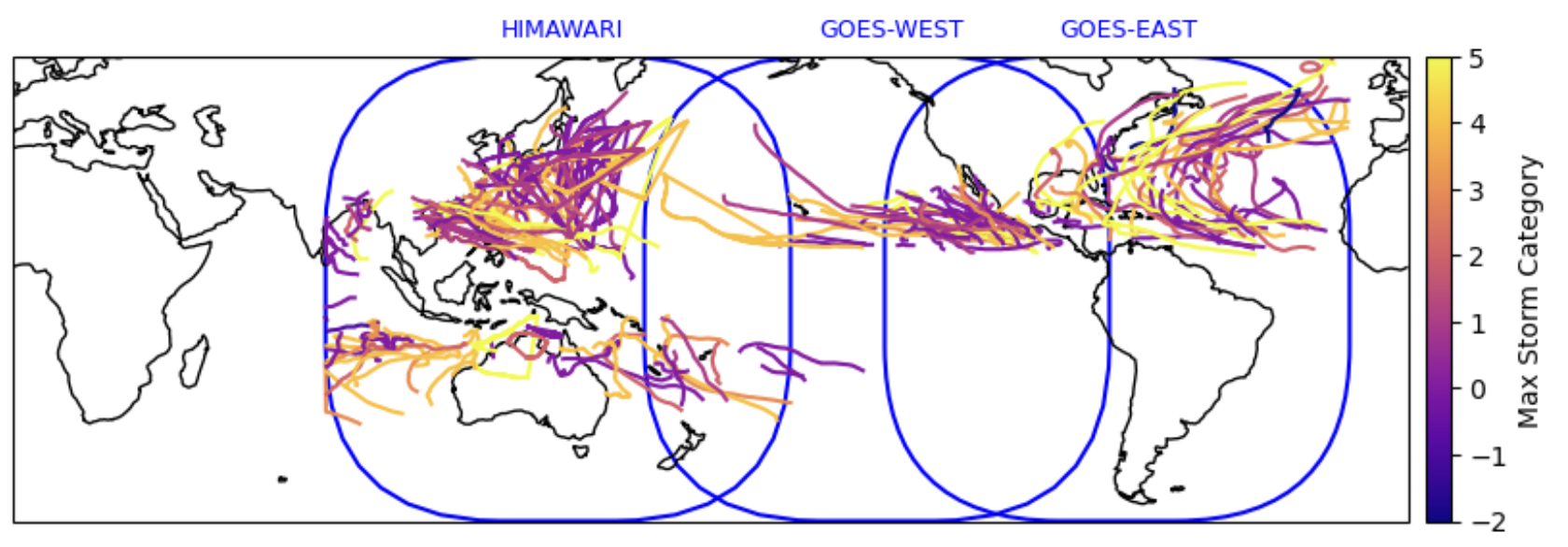}
    \caption{Field-of-view of the geostationary satellites used in this work. Future work will expand our approach to also include coverage from the Meteosat Second Generation (MSG) satellite to complete coverage over the Indian Ocean.}
    \label{fig:field-of-view}
\end{figure}

\section{Hyperparameters}
\label{app:hyperparams}
We use four-level residual U-Nets with channel widths $(32,64,128,256)$, GroupNorm, SiLU activations, and skip connections, operating on $192\times192$ centre crops. The MLP has hidden widths 128 and 64 with dropout 0.1, and no decoder is used. All models use AdamW with weight decay $10^{-4}$. U-Net and encoder-only models use batch size 4 and learning rate $2\times10^{-4}$, while the two-stage model uses batch size 16 and learning rate $3\times10^{-4}$. We reduce the learning rate on validation plateaus and use early stopping with patience 50.

\end{document}